\documentclass[11pt,a4paper]{article}
\usepackage[hyperref]{naaclhlt2019}
\usepackage{times}
\usepackage{etex}  
\usepackage{standalone}
\usepackage{latexsym}
\usepackage{url}
\usepackage{amsmath}
\usepackage{amssymb}
\usepackage{amsfonts}
\usepackage{graphicx}
\usepackage[skip=5pt]{caption}
\usepackage{subcaption}
\usepackage{array}
\usepackage{tabu}
\usepackage{makecell}
\usepackage{subcaption}
\usepackage{paralist}
\usepackage{cases}
\usepackage{diagbox}
\usepackage{enumitem}
\usepackage{soul}
\usepackage{multirow}
\usepackage{verbatim}
\usepackage{tabulary}
\usepackage{booktabs}
\usepackage[mathscr]{euscript}
\usepackage{mathtools}
\usepackage{amssymb}
\usepackage{algorithm}
\usepackage{algpseudocode}
\usepackage{stmaryrd}
\usepackage{amsthm}
\usepackage{tikz-dependency}
\usetikzlibrary{automata,decorations.markings,arrows,positioning,matrix,calc,patterns,angles,quotes,calc}
\usepackage{adjustbox}
\usepackage[toc,page]{appendix}
\usepackage{tabularx}
\usepackage{xspace}
\usepackage[toc,page]{appendix}

\aclfinalcopy 
\def\aclpaperid{1829}

\algrenewcommand{\algorithmiccomment}[1]{\leavevmode\hfill$\triangleright$ #1}

\usepackage{color}
\usepackage{bm}
\definecolor{orange}{rgb}{1,0.5,0}
\definecolor{mdgreen}{rgb}{0.05,0.6,0.05}
\definecolor{mdblue}{rgb}{0,0,0.7}
\definecolor{dkblue}{rgb}{0,0,0.5}
\definecolor{dkgray}{rgb}{0.3,0.3,0.3}
\definecolor{slate}{rgb}{0.25,0.25,0.4}
\definecolor{gray}{rgb}{0.5,0.5,0.5}
\definecolor{ltgray}{rgb}{0.7,0.7,0.7}
\definecolor{purple}{rgb}{0.7,0,1.0}
\definecolor{lavender}{rgb}{0.65,0.55,1.0}

\definecolor{mypurple}{RGB}{111,61,121}
\definecolor{myblue}{RGB}{46,88,180}
\definecolor{myred}{RGB}{181,68,106}
\definecolor{myyellow}{RGB}{204,143,55}

\newcommand{\ensuretext}[1]{#1}
\newcommand{\marker}[2]{\ensuremath{^{\textsc{#1}}_{\textsc{#2}}}}

\newcommand{\arkcomment}[3]{\ensuretext{\textcolor{#3}{[#1 #2]}}}
\renewcommand{\arkcomment}[3]{}  

\newcommand{\ankur}[1]{\arkcomment{\marker{A}{P}}{#1}{mdgreen}}

\newcommand{\hao}[1]{\arkcomment{\marker{H}{P}}{#1}{orange}}

\newcommand{\com}[1]{}
\newcommand{\resolved}[1]{}
\newcommand{\camready}[1]{}

\newcommand{\interalia}[1]{\citep[\emph{inter alia}]{#1}}

\newcommand{\softmax}{\operatorname{softmax}}

\newcolumntype{L}[1]{>{\raggedright\let\newline\\\arraybackslash\hspace{0pt}}m{#1}}
\newcolumntype{C}[1]{>{\centering\let\newline\\\arraybackslash\hspace{0pt}}m{#1}}
\newcolumntype{R}[1]{>{\raggedleft\let\newline\\\arraybackslash\hspace{0pt}}m{#1}}

\theoremstyle{definition}

\theoremstyle{remark}

\newcommand{\repo}{\url{https://homes.cs.washington.edu/~hapeng}}

\setul{1pt}{.4pt}

\DeclareCaptionFont{tiny}{\tiny}

\usepackage{url}
\usepackage[shortcuts]{extdash}  

\def\va{{\mathbf{a}}}
\def\vb{{\mathbf{b}}}
\def\vc{{\mathbf{c}}}

\def\vf{{\mathbf{f}}}
\def\vg{{\mathbf{g}}}
\def\vh{{\mathbf{h}}}

\def\vu{{\mathbf{u}}}
\def\vv{{\mathbf{v}}}
\def\vw{{\mathbf{w}}}
\def\vx{{\mathbf{x}}}
\def\vy{{\mathbf{y}}}
\def\vz{{\mathbf{z}}}

\def\mA{{\mathbf{A}}}
\def\mB{{\mathbf{B}}}
\def\mC{{\mathbf{C}}}

\def\mP{{\mathbf{P}}}
\def\mQ{{\mathbf{Q}}}

\def\mU{{\mathbf{U}}}
\def\mV{{\mathbf{V}}}

\newcommand{\rouge}{\textsc{Rouge}}
\newcommand{\shortrouge}{\textsc{Rg}}
\newcommand{\bleu}{\textsc{Bleu}}

\newcommand{\basic}{\textsc{Seq2seq}\xspace}
\newcommand{\an}{\textsc{AttExp}\xspace}
\newcommand{\ours}{\textsc{AdaDec}\xspace}
\newcommand{\both}{\textsc{AdaDec+AttExp}\xspace}

\title{Text Generation with Exemplar-based Adaptive Decoding}

\author{Hao Peng$^{\spadesuit\ast}$ \ 
Ankur P. Parikh$^\diamondsuit$  \ 
Manaal Faruqui$^\diamondsuit$ \ 
Bhuwan Dhingra$^{\clubsuit\ast}$ \  
Dipanjan Das$^\diamondsuit$ \\
$^\spadesuit$ Paul G. Allen School of Computer Science \& Engineering, University of Washington, Seattle, WA\\
$^\diamondsuit$ Google AI Language, New York, NY\\
$^\clubsuit$ School of Computer Science,
	Carnegie Mellon University, Pittsburgh, PA\\
	{\tt hapeng@cs.washington.edu,
	\{aparikh,mfaruqui,dipanjand\}@google.com,}\\
	{\tt bdhingra@andrew.cmu.edu}
}

\date{}

\begin{document}
\maketitle

\setlength{\abovedisplayskip}{6pt}
\setlength{\belowdisplayskip}{6pt}
\begin{abstract}
We propose a novel conditioned text generation model.
It draws inspiration from traditional template-based text generation techniques,
where the source provides the content (i.e., \textit{what to say}), 
and the template influences \textit{how to say it}.
Building on the successful encoder-decoder paradigm,
it first encodes the content representation from the given input text;
to produce the output,
it retrieves exemplar text from the training data as ``soft templates,''
which are then used to construct an exemplar-specific decoder.
We evaluate the proposed model on abstractive text
summarization and data-to-text generation.
Empirical results show that this model
achieves strong performance and outperforms comparable baselines.
\hao{change this argument after we have more results.}
\end{abstract}
\section{Introduction}
\let\svthefootnote\thefootnote
\let\thefootnote\relax\footnote{$\ast$ Work done during internship at Google.}
\addtocounter{footnote}{-1}\let\thefootnote\svthefootnote
Conditioned text generation is the essence of many
natural language processing (NLP) tasks, e.g.,
text summarization~\citep{mani1999advances}, machine translation~\citep{koehn2009statistical}, and data-to-text generation~\citep{kukich1983design,mckeown1992text, reiter1997building}.
In its common neural sequence-to-sequence formulation~\citep{sutskever2014sequence,cho2014learning},
an encoder-decoder architecture is used. 
The decoder generates the text autoregressively, token-by-token, 
conditioning on the feature representations encoded from the source, 
typically with attention~\citep{Bahdanau:2015} and copy mechanisms~\citep{gu2016incorporating,see2017get}. 
This paradigm is capable of generating fluent abstractive text,
but in an uncontrolled and sometimes unreliable way,
often producing degenerate outputs and favoring generic utterances~\cite{vinyals2015neural, li2015diversity}.

The encoder-decoder approach differs considerably from earlier template-based
methods~\interalia{becker2002practical,foster2004techniques,reiter2005choosing,gatt2009simplenlg},
where the source content is filled into the slots of a handcrafted template. 
These solutions offer higher generation precision compared to neural approaches~\cite{wiseman2017challenges},
but tend to lack the naturalness of neural systems, and are less scalable 
to open domain settings, where the number of required templates can be prohibitively large.

To sidestep the scalability problems with handcrafted templates, 
it has been proposed to use similar training samples as \textbf{exemplars},
to guide the decoding process~\interalia{gu2017search, guu2018generating, weston2018retrieve,pandey2018exemplar,cao2018retrieve}.%
\footnote{The term \textbf{exemplar}
	indicates a training instance used to help generation. 
	We aim to distinguish from ``templates,''
	since here no explicit slot-filling procedure is involved.}
In general, existing methods accomplish this by 
(a) using traditional information retrieval (IR) techniques
for exemplar extraction (e.g., TF-IDF), 
and then (b) concatenating the exemplar to the source as additional inputs, 
allowing the decoder to attend over and copy from both. 

We propose a different strategy for using exemplars. 
For motivation, Figure~\ref{fig:ex} shows a source-target pair together with its exemplar
from the Gigaword dataset~\citep{graff2003giga}.
The target is a summary of the source sentence,
and the exemplar is retrieved from the training set (\S\ref{subsec:model:retrieval}).%
\footnote{We use the \emph{training target} as the exemplar, 
	whose source is most similar to the current input.
	\S\ref{subsec:model:retrieval} describes the details.
}
There is word overlap between the exemplar and the desired output,
which would be easily captured by an attention/copy mechanism (e.g. \textit{Norway} and \textit{aid}).
Despite this, ideally, the model should also exploit the structural and stylistic aspects 
to produce an output with a similar sentence structure, even if the words are different.



\begin{figure}[tb]
	\fbox{
		\fontfamily{qcr}\selectfont
		\begin{minipage}{0.45\textwidth}
			\begin{small}
				{\color{myblue}\textbf{Source}}: Norway said Friday it would give Zimbabwe 40 million kroner (7.02 million dollars, 4.86 million euros) in aid to help the country deal with a lack of food and clean drinking water and a cholera outbreak. 
				
				{\color{myblue}\textbf{Exemplar}}:  Norway boosts earthquake aid to Pakistan.
				
				{\color{myblue}\textbf{Target}}: Norway grants aid of 4.86 million euros to Zimbabwe.
			\end{small}
		\end{minipage}
	}
	\caption{A source-target pair from Gigaword training set,
		 along with its exemplar.}
	\label{fig:ex}
\end{figure}

Indeed, in traditional templates,  the source is supposed to determine ``what to say,'' 
while the templates aim to address ``how to say it,'' 
reminiscent of the classical content selection and surface realization pipeline~\citep{reiter1997building}. 
For instance, an ideal template for this example might look as follows: \\ 
\texttt{\underline{\hspace{1cm}} grants aid of \underline{\hspace{1cm}} to \underline{\hspace{1cm}}} \\
\\
In the neural formulation, the ``how to say it'' aspect is primarily controlled by the decoder.

Inspired by the above intuition, we propose 
\textbf{exemplar-based adaptive decoding},
where a customized decoder is constructed for each exemplar.
This is achieved by 
letting the exemplars to directly influence decoder parameters
through a reparameterization step~(\S\ref{subsec:model:reparam}).
The adaptive decoder can be used as a drop-in replacement 
in the encoder-decoder architecture.
It offers the potential to better incorporate 
the exemplars' structural and stylistic aspects into decoding,
without excessive increase in the amount of parameters
or computational overhead.
\ankur{can we give some intuition as to why this is good?} 
\hao{does this seem okay? i'm trying to avoid overclaiming}

We empirically evaluate our approach on abstractive text summarization 
and data-to-text generation~(\S\ref{sec:experiments}), 
on which most of the recent efforts on exemplar-guided text generation have been studied. 
On three benchmark datasets, our approach outperforms
comparable baselines, and achieves performance competitive with the state of the art.
The proposed method can be applicable in many other conditioned text generation tasks. 
Our implementation is available at~\repo.


\section{Background}
\label{sec:background}
This section lays out the necessary
background and notations for further technical discussion.
We begin with conditioned text generation and the encoder-decoder framework~\citep{sutskever2014sequence,cho2014learning}.
In the interest of the notation clarity,
\S\ref{sec:model} will use an Elman network~(\citealp{Elman:1990})
as a running example for the decoder,
which is briefly reviewed in~\S\ref{sec:model}.
The proposed technique generalizes 
to other neural network architectures~(\S\ref{subsec:model:discussion}).



\paragraph{Conditioned text generation and the encoder-decoder architecture.}
\
Our discussion centers around 
conditioned text generation, i.e.,
the model aims to output the target $\vy = y_1 y_2\dots y_T$ given
the source input $\vx = x_1 x_2\dots x_S$,
both of which are sequences of tokens.
Each token $x_i$, $y_i$ takes one value from 
a vocabulary $\mathcal{V}$.
$\vx$ and $\vy$ could vary depending on the tasks,
e.g., they will respectively be articles and summaries for text summarization;
and for data-to-text generation, 
$\vx$ would be structured data, which can sometimes be linearized~\interalia{lebret2016wikibio,wiseman2018learning}, and $\vy$ is the output text.
We aim to learn a (parameterized) conditional distribution
of the target text $\mathbf{y}$ given the source $\mathbf{x}$,
\begin{align}
\label{eq:seq2seq}
p\bigl(\vy \mid \vx\bigr) = \prod_{t=1}^{T} p\bigl(y_t \mid \vy_{<t}, \vx\bigr),
\end{align}
where $\vy_{<t}=y_1\dots y_{t-1}$ is the prefix of $\vy$ up to the $(t-1)^\text{th}$ token (inclusive).

The probability of each target token is usually estimated with a $\softmax$~function:
\begin{align}
\label{eq:softmax}
p\left(y_t \mid \vy_{<t}, \vx\right) 
= \frac{\exp{ \vh_{t-1}^\top\vw_{y_{t}}}}{\sum_{y} \exp{ \vh_{t-1}^\top\vw_{y}}}.
\end{align}
$\vw_{y}$ denotes a learned vector for token $y\in\mathcal{V}$.
$\vh_{t-1}$ depends on $\vy_{<t}$ and $\vx$, and is computed by a function which we will describe soon.

A typical implementation choice for computing $\vh_{t}$ 
is the encoder-decoder architecture~\citep{sutskever2014sequence}.
More specifically, 
an \textbf{encoder} $\vg_{\bm{\theta}}$ first gathers the feature representations
from the source $\vx$;
then a \textbf{decoder} $\vf_{\bm{\phi}}$ is used to compute the $\vh_t$ feature vectors:
\begin{align}
\vh_t = \vf_{\bm{\phi}}\Bigl(y_t, \vh_{t-1},\vg_{\bm{\theta}}(\vx)\Bigr).
\end{align}
$\bm{\theta}$ and $\bm{\phi}$ are, respectively, the collections of parameters
for the encoder and the decoder, both of which
can be implemented 
as recurrent neural networks (RNNs) such as LSTMs~\cite{Hochreiter:1997} or GRUs~\citep{cho2014learning},
or the transformer~\citep{vaswani2017attention}.
In~\citet{sutskever2014sequence},
the dependence of $\vf_{\bm{\phi}}$ on $\vg_{\bm{\theta}}$ is made by using the last 
hidden state of the encoder as the initial state of the decoder.
Such dependence can be further supplemented 
with attention~\citep{Bahdanau:2015} and copy mechanisms~\citep{gu2016incorporating,see2017get}, 
as we will do in this work.

\S\ref{sec:model} introduces how we 
use exemplars to inform decoding, by
dynamically constructing the decoder's parameters~$\bm{\phi}$.
For the notation clarity, we will use the Elman network as a running example, reviewed below.

\paragraph{Elman networks.}
Given input sequence $\mathbf{x}$,
an Elman network~\citep{Elman:1990} computes the hidden state at time step $t$
from the previous one and the current input token by
\begin{align}
\label{eq:elman}
\mathbf{h}_t=\tanh\bigl( \mP\mathbf{h}_{t-1} + \mQ\vv_{t}\bigr),
\end{align}
where $\mP$ and $\mQ$ are learned $d\times d$ 
parameter matrices (with $d$ being the hidden dimension),
and $\vv_t$ is the embedding vector for token $x_t$.
We omit the bias term for clarity.

\section{Method}
\label{sec:model}
This section introduces the proposed method in detail.
Our aim is to use exemplars to inform
the decoding procedure (i.e., \emph{how to say it}).
To accomplish this, we reparameterize
the decoder's parameters with weighted linear sums,
where the coefficients are determined by an exemplar. 
The decoder is \textbf{adaptive},
in the sense that its parameters vary 
according to the exemplars.
The adaptive decoder can be used as a
drop-in replacement in the encoder-decoder architecture.
Before going into details,
let us first overview the high-level generation procedure
of our model.
Given source text $\vx$, the model generates an output as follows:
 \begin{compactitem}
 \item[1.] Run a standard encoder to gather the content representations $\vg_{\bm{\theta}}(\vx)$ from the source.
 \item[2.] Retrieve its exemplar $\vz_{\vx}$, 
 and compute exemplar-specific coefficients~(\S\ref{subsec:model:retrieval}).
 \item[3.] Construct the adaptive decoder parameters $\bm{\phi}$~(\S\ref{subsec:model:reparam}), 
 using the coefficients computed at step 2.
 Then the output is generated by applying the adaptive decoder followed by a $\softmax$, 
 just as in any other encoder-decoder architecture.
 \end{compactitem}
Aiming for a smoother transition, 
we will first describe step 3 in \S\ref{subsec:model:reparam},
and then go back to discuss step 2 in \S\ref{subsec:model:retrieval}.
For clarity, we shall assume 
that the decoder is implemented as
an Elman network~(\citealp{Elman:1990};~Equation~\ref{eq:elman}).
The proposed technique generalizes
to other neural network architectures, as we will discuss later in \S\ref{subsec:model:discussion}.

\subsection{Reparameterizing the RNN Decoder}
\label{subsec:model:reparam}
At its core, the exemplar-specific adaptive decoder
involves a reparameterization step, which we now 
describe. 
We focus on the parameters of the Elman network decoder, i.e., 
$\mP$ and $\mQ$ in~Equation~\ref{eq:elman}.

\paragraph{Parameter construction with linear sums.}
\
We aim to reparameterize the pair of matrices $(\mP, \mQ)$,
in a way that they are influenced by the exemplars.

Let us first consider an extreme case, 
where one assigns a different pair of parameter matrices
to each exemplar, \emph{without} any sharing.
This leads to an unreasonably large amount of parameters,
which are difficult to estimate reliably.\footnote{
	The amount of parameters grows linearly with the number of possible exemplars,
	which, as we will soon discuss in \S\ref{subsec:model:retrieval}, can be as large as the training set.}

We instead
construct $\mP$ and $\mQ$ from a set of predefined 
parameters matrices. 
Take $\mP$ for example,
it is computed as the weighted sum of $\mP_i$ matrices:
\begin{align}
\label{eq:sum}
\mP = \sum_{i=1}^{r}\lambda_i\mP_i,
\end{align}
where $\mP_i\in\mathbb{R}^{d\times d}$, with $d$ being the size of the hidden states.
$r$ is a hyperparameter, determining the number of $\mP_i$ matrices to use.%
\footnote{Instead of choosing $r$ empirically, we set it equal to $d$ in the experiments. Please see the end of \S\ref{subsec:model:reparam}
	for a related discussion.} 
The summation is weighted by the coefficients $\lambda_i$,
which are computed from the exemplar 
$\vz_{\vx}$.
For clarity, the dependence of both $\mP$ and $\lambda_i$ 
on $\vz_{\vx}$ is suppressed when the context is clear.

Equation~\ref{eq:sum} constructs the
decoder's parameter matrix $\mP$ using a linear combination
of $\{\mP_i\}_{i=1}^{r}$.
The exemplar informs this procedure through the coefficients $\lambda_i$'s,
the detailed computation of which is deferred to \S\ref{subsec:model:retrieval}.
The other matrix $\mQ$ can be similarly constructed by $\mQ = \sum_{i}\lambda_i\mQ_i$.

\paragraph{Rank-1 constraints.}
\
In the above formulation, the number of parameters is
still $r$ times more than a standard Elman network,
which can lead to overfitting with a limited amount of training data. 
Besides, it would be more interesting to compare the adaptive decoder
to a standard RNN under a comparable parameter budget.
Therefore we want to further limit the amount of parameters.
This can be achieved by forcing the ranks of $\mP_i$ and $\mQ_i$ to be 1,
since it then takes $2d$ parameters to form each of them, instead of $d^2$.
More formally, we upper-bound their ranks by construction:
\begin{align}
\label{eq:rank1}
\mP_i = \mathbf{u}_i^{(p)}\,\otimes\,\mathbf{v}_i^{(p)}.
\end{align}
$\mathbf{a}\otimes\mathbf{b}=\mathbf{a}\mathbf{b}^\top$ 
denotes the outer product of two vectors;
$\vu_i^{(p)}$ and $\vv_i^{(p)}$ are learned $d$-dimensional vectors.
Each $\mQ_i$ can be similarly constructed by a separate set of 
vectors $\mQ_i = \vu_i^{(q)} \otimes \vv_i^{(q)}$.

Let $\mU_p, \mV_p\in \mathbb{R}^{d\times r}$ denote the stack
of $\vu_i^{(p)}$, $\vv_i^{(p)}$ vectors, i.e.,
\begin{subequations}
	\begin{align}
	\mU_p&=\Bigl[\vu_1^{(p)}, \dots,\vu_r^{(p)} \Bigr], \\ 
	\mV_p&=\Bigl[\vv_1^{(p)}, \dots,\vv_r^{(p)}\Bigr].
	\end{align}
\end{subequations}
Equations~\ref{eq:sum} and~\ref{eq:rank1} can be compactly
written as 
\begin{align}
\label{eq:p_compact}
\mP&=\mU_p\;\bm{\Lambda}\;\mV_p^\top.
\end{align}
where $\bm{\Lambda}$ is the diagonal matrix built from the $r$-dimensional 
coefficient vector $\bm{\lambda}=[\lambda_1, \dots,\lambda_r]^\top$:
\begin{align}
\bm{\Lambda}=
\operatorname{diag}(\bm{\lambda})=
\begin{bmatrix}
\lambda_1 & &\\
& \ddots &\\
& & \lambda_r
\end{bmatrix}.
\end{align}
The construction of $\mQ$ is similar, 
but with a different set of parameters matrices $\mU_q$ and $\mV_q$:%
\footnote{The bias term in the Elman network $\mathbf{b}$ can be constructed as $\vb=\mB\bm{\lambda}$, with $\mathbf{B}$ being a learned $d\times r$ matrix.}
\begin{align}
\label{eq:q_compact}
\mathbf{Q}&=\mU_q\;\bm{\Lambda}\;\mV_q^\top.
\end{align}
Note that, despite their similarities to SVD at a first glance,
Equations~\ref{eq:p_compact} and~\ref{eq:q_compact} are not performing matrix factorization.
Rather, we are learning $\{\mU_p,\mV_p,\mU_q,\mV_q\}$ directly;
$\mP$, $\mQ$, $\{\mP_i\}$, and $\{\mQ_i\}$ are never explicitly instantiated~\citep{peng2017deep,peng2018learning}.

To summarize, we reparameterize $\mP$ and $\mQ$
as interpolations of rank-1 matrices.
By the fact that~$\operatorname{rank}(\mA+\mB)\leq\operatorname{rank}(\mA)+\operatorname{rank}(\mB)$,
the ranks of $\mP$ and $\mQ$ are upper-bounded by $r$.
As pointed out by \citet{krueger2016regularizing}, the parameter matrices of a trained RNN
tend to have full rank. 
Therefore, in the experiments, we set $r$ equal to the hidden size $d$,
aiming to allow the adaptive decoder to use full-rank matrices in the recurrent computation.
Yet,
if one holds \textit{a priori} beliefs that the matrices should have lower ranks,
using $r<d$ could be desirable.
When $r=d$, an adaptive RNN constructed by the above approach
has $4d^2$ parameters, 
which is comparable to the $2d^2$ parameters in a standard Elman network.%
\footnote{This does not include the bias term, which contributes
	additional $d^2$ parameters to the former, and $d$ to the latter.}

\subsection{Incorporating Exemplars}
\label{subsec:model:retrieval}
We now discuss the computation of coefficients $\bm{\lambda}$,
through which the exemplars inform
the decoder construction~(Equations~\ref{eq:p_compact} and~\ref{eq:q_compact}).
Before detailing the neural network architecture, 
we begin by describing the exemplar retrieval procedure.

\paragraph{Retrieving exemplars $\vz_{\vx}$.}
\
Intuitively, similar source texts should hold similar targets.
Therefore, given source input $\vx$, we use 
the \emph{training target} as its exemplar $\vz_{\vx}$, whose source is most similar to $\vx$.%
\footnote{
	The source of an exemplar is only used in the retrieval 
	and never fed into the encoder-decoder model.
	For a training instance, we additionally 
	disallow using its own target as the exemplar.}
To compute the similarities between source texts,
we use bag-of-words (BOW) features and cosine similarity.
We extract the top-1 exemplar for each instance.
This step is part of the pre-processing, and
we do \emph{not} change the exemplars as the training proceeds.

There are, of course, many other strategies to get the exemplars,
e.g., using handcrafted or heuristically created hard templates~\interalia{reiter2005choosing, becker2002practical,foster2004techniques},
randomly sampling \emph{multiple} training instances~\citep{guu2018generating},
or learning a neural reranker~\citep{cao2018retrieve}.
Using more sophistically extracted exemplars
is definitelly interesting to explore,
which we defer to future work.

 
\paragraph{Computing coefficients.}
Next we describe the computation of $\bm{\lambda}$, 
the $r$-dimensional coefficient vector,
which is used to construct the adaptive decoder~(Equations~\ref{eq:p_compact} and~\ref{eq:q_compact}).

\hao{i don't think the following is motivated well enough}
Intuitively, the rank-1 matrices~($\mP_i$'s and $\mQ_i$'s in Equation~\ref{eq:rank1} and thereafter)
can be seen as capturing different aspects 
of the generated text.
And $\bm{\lambda}$ determines how much each of them
contributes to the adaptive decoder construction.
A natural choice to calculate $\bm{\lambda}$ 
is to use the similarities between the exemplar and
each of the aspects.

To accomplish this, 
we run a RNN encoder over $\vz_{\vx}$,
and use the last hidden state as its vector representation $\va$.%
\footnote{For clarity, the dependence of $\va$ on the exemplar $\vz_{\vx}$
	is suppressed, just as $\bm{\lambda}$.}
We further associate each $(\mP_i, \mQ_i)$ pair
with a learned vector $\vc_i$;
and then $\lambda_i$ is computed as the similarity 
between $\va$ and $\vc_i$,
using an inner product $\lambda_i=\va^\top\vc_i$.
More compactly,
\begin{align}
\label{eq:lambda}
\bm{\lambda} = \mC\va,
\end{align}
with $\mC=[\vc_1,\dots,\vc_r]^\top$.

Closing this section, Algorithm~\ref{algo}
summarizes the procedure to construct an adaptive decoder.
\begin{algorithm}[tb]
	\centering
	\caption{Adaptive decoder construction.}
	\begin{algorithmic}[1]
		\Procedure{}{$\mathbf{x}$}
		\State Retrieve the exemplar $\vz_{\vx}$ 
		\Comment\S\ref{subsec:model:retrieval}
		\State Compute $\vz_{\vx}$'s representation $\va$
		\Comment\S\ref{subsec:model:retrieval}
		\State Compute coefficients $\bm{\lambda}$
		\Comment Eq.\ref{eq:lambda}
		\State Construct the decoder $\vf_{\bm{\phi}}$
		\Comment Eqs.\ref{eq:p_compact}, \ref{eq:q_compact}
		\EndProcedure
	\end{algorithmic}
	\label{algo}
\end{algorithm}

\subsection{Discussion.}
\label{subsec:model:discussion}
Although we've based our discussion
on Elman networks so far,
it is straightforward to apply this method
to its gated variants~\interalia{Hochreiter:1997,cho2014learning},
and other quasi-/non-recurrent neural architectures~\interalia{Bradbury:2017,vaswani2017attention,peng2018rational}.
Throughout the experiments, we will be using an adaptive LSTM decoder~(\S\ref{sec:experiments}).
As a drop-in replacement in the encoder-decoder architecture,
it introduces a reasonable amount of 
additional parameters and computational overhead,
especially when one uses a small encoder for the exemplar
(i.e., the sizes of the $\vc_i$ vectors in Equation~\ref{eq:lambda} are small).
It can benefit from the highly-optimized GPU
implementations, e.g., CuDNN,
since it uses the same recurrent computation as a standard nonadaptive RNN.

In addition to the neural networks,
the adaptive decoder requires access to the full training set due to the retrieval step.
In this sense it is \emph{semi-parametric}.%
\footnote{
	Nothing prohibits adaptively constructing
	other components of the model, e.g.,
	the encoder $\vg_{\bm{\theta}}$.
	Yet, our motivation is to use exemplars to inform \emph{how to say it},
	which is primarily determined by the decoder (in contrast,
	the encoder relates more to selecting the content).
}
The idea to dynamically construct the parameters
is inspired by Hypernetworks~\citep{ha2016hypernetworks} and earlier works therein.
It proves successful in tasks such as classification~\citep{xu2016dynamic,liu2017dynamic} 
and machine translation~\citep{platanios2018contextual}.
Many recent template-based generation models
include the exemplars as \emph{content} in addition to the source, 
and allow the decoder to attend over and copy from both~\interalia{gu2017search,guu2018generating, weston2018retrieve,pandey2018exemplar,cao2018retrieve}.
We compare to this approach in the experiments,
and show that our model offers favorable performance,
and that they can potentially be combined to achieve further improvements.

\begin{table}[tb]
	\center
	\begin{tabulary}{.47\textwidth}{@{}l  l rrr@{}}
		\toprule
		& &\bf{NYT} & \bf{Giga} & \bf{Wikibio}\\
		\midrule
		
		\multirow{3}{*}{\# inst.}
		& Train & 92K & 3.8M & 583K\\
		& Dev. & 9K & 190K & 73K\\
		& Test & 9,706 & 1,951 & 73K\\
		
		\midrule[.03em]
		
		\multirow{2}{*}{Avg. len.}
		& Src. & 939.0 & 31.4 & N/A\\
		& Tgt. & 48.6 & 8.2 & 26.0\\
		
		\bottomrule
	\end{tabulary}
	\caption{Number of instances and average text lengths for the datasets used in the experiments.
		The lengths are averaged over training instances.}
	\label{tab:data} 
\end{table}

\section{Experiments}
\label{sec:experiments}
This section empirically evaluates the proposed model
on two sets of text generation tasks:
abstractive summarization~(\S\ref{subsec:experiments:summuarization})
and data-to-text generation~(\S\ref{subsec:experiments:data2text}).
Before heading into the experimental details,
we first describe the architectures of the compared
models in \S\ref{subsec:experiments:models}.

\hao{about the title below,
	this section also describes our model,
	so it's not only about baselines}
\subsection{Compared Models}
\label{subsec:experiments:models}
In addition to previous works,
we compare to the following baselines,
aiming to control for confounding factors due to detailed implementation choices.

\begin{compactitem}
     \item\basic. The encoder-decoder 
     architecture enhanced 
     with attention and copy
     mechanisms.
     The encoder is implemented with a bi-directional LSTM~(BiLSTM; \citealp{Hochreiter:1997,shcuster1997bidirectional,graves2012supervised}), 
     and the decoder a uni-directional one.
     We tie the input embeddings of both the encoder and the decoder,
     as well as the $\softmax$ weights~\citep{press2017using}.
     We use beam search during evaluation, with length penalty~\citep{wu2016google}.
     
     \item\an. It is based on \basic.
     It encodes, attends over, and copies from the exemplars,
     in addition to the source inputs.
   
\end{compactitem}
Our model using the adaptive decoder (\ours)
closely builds upon \basic.
It uses a dynamically constructed LSTM decoder,
and \emph{does not} use attention or copy mechanisms over the encoded exemplars.
The extracted exemplars are the same as those used by \an.
To ensure fair comparisons,
we use comparable training procedures and regularization techniques
for the above models.
The readers are referred 
to the appendix for further details such as hyperparameters.

\subsection{Text Summarization}
\label{subsec:experiments:summuarization}
\paragraph{Datasets.}
\
We empirically evaluate our model on two benchmark text summarization datasets:
\begin{compactitem}
	\item Annotated Gigaword corpus~(Gigaword; \citealp{graff2003giga,napoles2012giga}). 
	Gigaword contains news articles sourced from various 
	news services over the last two decades.
	To produce the dataset, we follow the split and preprocessing by~\citet{rush2015neural},
	and pair the first sentences and the headlines in the news articles.
	It results in a 3.8M/190K/1,951 train/dev./test split. 
	The average lengths of the source and target texts are 31.4 and 8.2, respectively.
     \item New York Times Annotated Corpus (NYT;~\citealp{sandaus2018nyt}). 
     It contains news articles published between 1996 and 2007 by New York Times.
     We use the split and preprocessing by~\citet{durrett2016learning}.%
     \footnote{\url{https://github.com/gregdurrett/berkeley-doc-summarizer}.}
     Following their effort, we evaluate on a smaller portion of the test set,
     where the gold summaries are longer than 50 tokens.
     We further randomly sample 9,000 instances from the training data
     for validation, resulting in a 91,834/9,000/3,452 train/dev./test split. 
     Compared to Gigaword, the inputs and targets in NYT are much longer (averaging 939.0 and 48.6, respectively).
     
\end{compactitem}
Table~\ref{tab:data} summarizes some statistics of the datasets.
We note that some recent works use a different split of the NYT corpus~\citep{paulus2018deep,gehrmann2018bottom}, and thus are not comparable to the models in Table~\ref{tab:res:nyt}.
We decide to use the one by~\citet{durrett2016learning}
because their preprocessing script is publicly available.

For both datasets, we apply byte-paired encoding (BPE;~\citealp{sennrich2016bpe}),
which proves to improve the generation
of proper nouns~\citep{fan2018contollable}.

\begin{table}[tb]
	\centering
	\begin{tabulary}{0.47\textwidth}{@{}l  c@{\hskip 0.25cm}c@{\hskip 0.25cm}c@{}} 
		
		\toprule
		
		\textbf{Model}
		& \textbf{\shortrouge-1}
		& \textbf{\shortrouge-2}
		& \textbf{\shortrouge-L}\\
		\midrule
		
		Open-NMT & 35.0 & 16.6 & 32.4\\
		$^\dagger$\citealp{cao2018retrieve} (\textsc{Basic}) & 36.0 & 17.1 & 33.2\\
		$^\dagger$\citealp{cao2018retrieve} (\textsc{Full}) & 37.0 & \textbf{19.0} & 34.5\\
		$^\star$\citet{cao2017faithful} & \textbf{37.3} & 17.6 & 34.2\\
		\midrule[.03em]
		This work~(\basic) & 35.8 & 17.5 & 33.5\\
		$^\dagger$This work~(\an) & 36.0 & 17.7 & 33.1 \\
		\midrule[.03em]
		$^\dagger$This work~(\ours) & \textbf{37.3} & 18.5 & \textbf{34.7} \\
		
		\bottomrule
		
	\end{tabulary}
	\caption{Text summarization performance in \rouge~$F_1$ scores~(dubbed as \shortrouge-X)
		on Gigaword test set~(\S\ref{subsec:experiments:summuarization}).
		$\dagger$ denotes the models using retrieved exemplars,
		while~$\star$ uses handcrafted features.
		Bold font indicates best performance.
		Open-NMT numbers are taken from~\citet{cao2018retrieve}.}
	\label{tab:res:giga}
\end{table}

\paragraph{Empirical results.}
Table~\ref{tab:res:giga} compares the models on Gigaword test set
in \rouge~$F_1$~\citep{lin2004rouge}.\footnote{Version 1.5.5 of the official script.}


By using adaptive decoders, our model (\ours)
improves over \basic by more than 1.1 \rouge~scores.
\citet{cao2017faithful} and 
the \textsc{Full} model by~\citet{cao2018retrieve}
hold the best published results.
The former uses extensive handcrafted features
and relies on external information extraction and syntactic parsing systems;
while the latter uses additional encoding, attention and copy mechanisms over the exemplars
extracted using a novel neural reranker.
\ours achieves better or comparable performance to the state-of-the-art models,
\emph{without} using any handcrafted features or reranking techniques.
The \textsc{Basic} model by~\citet{cao2018retrieve}
ablates the reranking component from their \textsc{Full} model,
and uses the top exemplar retrieved by the IR system. 
Therefore it is a more comparable baseline to ours.
\ours outperforms it by more than 1.3 \rouge~scores.
Surprisingly, we do not observe interesting improvements by \an over the sequence-to-sequence baseline.
We believe that our model can benefit from better extracted exemplars by, e.g., applying a
reranking system. Such exploration is deferred to future work.

The NYT experimental results are summarized in Table~\ref{tab:res:nyt}.
We follow previous works
and report limited-length \rouge~\emph{recall} values.\footnote{
Following
\citet{durrett2016learning} and \citet{paulus2018deep},
we truncate the predictions to the lengths
	of the gold summaries,
	and evaluate \rouge~recall,
	instead of $F_1$ on full-length predictions.}
\citet{durrett2016learning}~is an extractive model, 
and \citet{paulus2018deep} an abstractive approach based on reinforcement learning.
Our \ours model outperforms both.
We observe similar trends when comparing \ours
to the \basic and \an baselines,
with the exception that \an \emph{does} improve over \basic.

\begin{table}[tb]
	\centering
	\begin{tabulary}{0.47\textwidth}{@{}l @{} cc@{}} 
		
		\toprule
		
		\textbf{Model}
		& \textbf{\rouge-1}
		& \textbf{\rouge-2}\\
		\midrule
		
		\citet{durrett2016learning} & 42.2 & 24.9\\
		\citet{paulus2018deep} & 42.9 & 26.0 \\
		\midrule[.03em]
		This work (\basic)  & 41.9 & 25.1 \\
		$^\dagger$This work (\an)  & 42.5 & 25.7 \\
		\midrule[.03em]
		$^\dagger$This work (\ours) & \textbf{43.2} & \textbf{26.4}\\
		
		\bottomrule
		
	\end{tabulary}
	\caption{NYT text summarization test performance
	in \rouge~\emph{recall} values.
	This is a smaller portion of the original test data,
	after filtering out instances with summaries shorter than 50 tokens~(\S\ref{subsec:experiments:summuarization};~\citealp{durrett2016learning}).
	$\dagger$ denotes the models using retrieved exemplars,
	and bold font indicates best performance.}
	\label{tab:res:nyt}
\end{table}

\newcommand{\cell}[1]{\begin{tabular}{@{}l@{}}\centering#1\end{tabular}}

\subsection{Data-to-text Generation}
\label{subsec:experiments:data2text}
\begin{figure}[!tb]
	\centering
	\begin{subfigure}[b]{.9\columnwidth}
		\centering
		\includegraphics[clip,trim=5cm 0cm 5cm 0cm, width=.9\columnwidth]{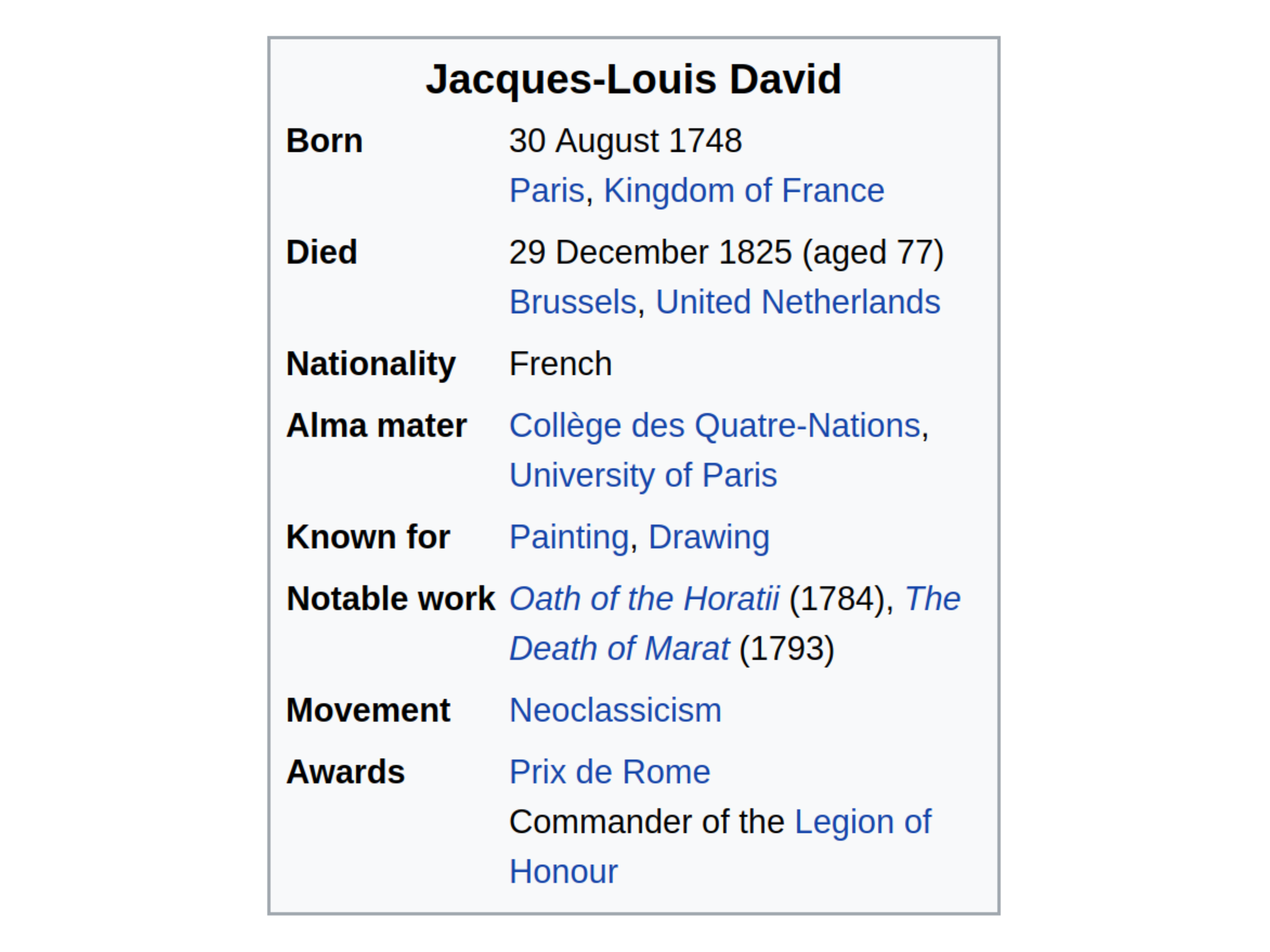}
		\vspace{-.4cm}
	\end{subfigure}
	
	\rule{.9\columnwidth}{.02cm}
	\vspace{.2cm}
	
	\begin{subfigure}[b]{.9\columnwidth}
		\flushleft
		Jacques-Louis David~(30 August 1748~-- 29 December 1825) was a French painter in the Neoclassical style.
		\vspace{-.2cm}
	\end{subfigure}
	
	\rule{.9\columnwidth}{.02cm}
	\caption{A training instance from the Wikibio dataset. 
		It consists of a collections of records
		for Jacques-Louis David~(top), and a piece of textual description~(bottom).}
	\label{fig:wikibio_example}
\end{figure}

Data-to-text generation aims to generate textual descriptions of structured data,
which can be seen as a table consisting of a collection of records~\citep{liang2009learning}.
For a given entity,
each record is an~\textit{(attribute, value)}~tuple.
Figure~\ref{fig:wikibio_example}
shows an example for entity \textit{Jacques-Louis David}.
The table specifies the entity's properties with tuples
\textit{(born, 30 August 1748)}, \textit{(nationality, French)},
and so forth.
The table is paired with a description,
which the model is supposed to generate
using the table as input.
We refer the readers to~\citet{lebret2016wikibio} for 
more details about the task.

\paragraph{Dataset and implementation details.}
\
We use the Wikibio dataset~\citep{lebret2016wikibio}.
It is automatically constructed by
pairing the tables and the opening sentences
of biography articles from English Wikipedia.
We follow the split and preprocessing provided along with
the dataset, with around 583K/73K/73K train/dev./test instances.
Following~\citet{lebret2016wikibio}, we linearize the tables,
such that we can conveniently train 
the sequence-to-sequence style models described in \S\ref{subsec:experiments:models}.
Table~\ref{tab:data} summarizes some statistics of the dataset.

In contrast to the text summarization experiment~(\S\ref{subsec:experiments:summuarization}),
we do \textit{not} apply BPE here.
Further, the word embeddings are initialized with
\texttt{GloVe}~(\citealp{pennington2014glove}; fixed during training), 
and \emph{not} tied with the $\softmax$ weights.
In addition to the models introduced in \S\ref{subsec:experiments:models},
we additionally compare to \both,
aiming to study whether the adaptive decoder
can further benefit from 
attention and copy mechanisms over the exemplars.

\paragraph{Empirical results.}
\
Following \citet{liu2017table}
we report~\rouge-4 and~\bleu~scores~\citep{papineni2002bleu}.%
\footnote{We use the script by~\citet{lin2004rouge} to calculate the \rouge~score,
	and the mteval script for \bleu: 
	\url{https://github.com/moses-smt/mosesdecoder/blob/master/scripts/generic/mteval-v13a.pl}.}
Table~\ref{tab:res:wikibio}
summarizes the data-to-text generation results on 
the Wikibio test set.
Overall, we observe similar trends
to those in the summarization experiment~(\S\ref{subsec:experiments:summuarization}):
by attending over and copying from the exemplars,
\an~improves upon the \basic baseline by around 0.6 absolute scores.
Also utilizing exemplar information, our \ours~model outperforms \basic by a larger margin:
1.3 for \rouge-4 and 1.1 for \bleu.
We further study whether we can get further improvements
by combining both.
\both~achieves around 0.5 absolute improvements over \ours,
less than those by \an over \basic.
This provides evidence that, to some extend,
the ways \an~and \ours incorporate exemplar information
might be complementary.
\citet{wiseman2018learning} is a template-motivated model
based on a semi-Markov model.
\citet{liu2017table}
hold the current state-of-the-art results.
They encode the table structures by 
using (a) position and filed embeddings,
and (b) structure-aware attention and gating techniques.
These techniques are beyond the scope of this work,
which focuses mainly on the decoding end.

 \begin{table}[tb]
 	\center
 	\begin{tabulary}{0.47\textwidth}{@{}l @{\hskip 0.25cm} c@{\hskip 0.4cm}c@{}} 
		
 		\toprule
		
 		\bf{Model}
 		& \bf{\shortrouge-4}
 		& \bf{\bleu}\\
 		\midrule
 	    \citet{wiseman2018learning} & 38.6 & 34.8  \\
 	    \citet{liu2017table} & 41.7 & 44.7 \\
 		\midrule[.03em]
 		This work (\basic) & 39.3 & 42.5\\
 		$^\dagger$This work (\an) & 40.0 & 43.1 \\
 		\midrule[.03em]
 		$^\dagger$This work (\ours) & 40.6 & 43.6 \\
 		$^\dagger$This work (\both) & 41.1 & 44.1 \\
 		
 		\bottomrule
 	\end{tabulary}
 	\caption{Data-to-text generation performance in \rouge-4 and \bleu~on the Wikibio test
 		set~(\S\ref{subsec:experiments:data2text}).
 			$\dagger$ indicates the models using retrieved exemplars.}
 	\label{tab:res:wikibio}
 \end{table}

\section{Analysis}
\label{sec:analysis}
We now qualitatively evaluate our model, by studying how 
its outputs are affected by using different exemplars.
Figure~\ref{fig:analysis} shows two randomly sampled Gigaword development instances.
It compares the outputs by 
\ours~(i.e., without attention/copy over exemplars; \S\ref{subsec:experiments:models})
when receiving different exemplars,
controlling for the same source inputs.
In each example, 
{\fontfamily{qcr}\selectfont Exemplar 1}
is retrieved by the system (i.e., a training target; \S\ref{subsec:model:retrieval});
while the remaining ones are produced by the authors,
by modifying the first one in styles 
and sometimes introducing distractions in the content.


In the top example,
the model includes \textit{people} into the subject (\textit{Three} vs. \textit{Three people})
 under the influence by {\fontfamily{qcr}\selectfont Exemplar 2};
{\fontfamily{qcr}\selectfont Exemplar 3}
changes the tense and adds some distraction by
changing the place from \textit{Britain} to \textit{Canada}.
The model follows the tense switch,
but gets confused by the distraction,
and decides to let a train in southern Europe collide into North America,
which it should not.
Looking at the bottom example, the model in general follows the 
exemplar in using noun adjuncts or prepositional phrases
(e.g., \textit{new home sales} vs. \textit{sales of new homes}), except the first one.
Perhaps confused by the distraction in {\fontfamily{qcr}\selectfont Exemplar 3},
the model makes a judgment on the specific amount of growth,
but gets it wrong.

\begin{figure}[h]
	\fbox{
		\fontfamily{qcr}\selectfont
		\begin{minipage}{0.45\textwidth}
			\begin{small}
				{\color{myblue}\textbf{Source}}: A Portuguese train derailed in the northern region of Oporto on Wednesday, killing three people...\\
				
				{\color{myblue}\textbf{Exemplar 1}}:  Two die in a Britain train collision.\\
				{\color{myblue}\textbf{Output 1}}: Three killed in Portuguese train derailment. \\
				
				{\color{myblue}\textbf{Exemplar 2}}:  Two~{\color{mypurple}\textbf{people were killed}}~in Britain train collision.\\
				{\color{myblue}\textbf{Output 2}}: Three {\color{myyellow} \textbf{\emph{people}}} killed in Portuguese train derailment. \\
				
				{\color{myblue}\textbf{Exemplar 3}}:  A train collision in {\color{mypurple}\textbf{Canada killed two people}}.\\
				{\color{myblue}\textbf{Output 3}}: Portuguese train derails in {\color{myyellow} \textbf{\emph{northern Mexico killing three}}}. 
			\end{small}
		\end{minipage}
	}
	
		\fbox{
			\fontfamily{qcr}\selectfont
			\begin{minipage}{0.45\textwidth}
				\begin{small}
					{\color{myblue}\textbf{Source}}: Sales of new homes in the U.S. increased by 11.8 percent in May, the biggest gain in 26 years... \\
					
					{\color{myblue}\textbf{Exemplar 1}}:  U.S. sales of new homes up strongly in March.\\
					{\color{myblue}\textbf{Output 1}}: US new home sales rise 11.8 percent in May. \\
					
					{\color{myblue}\textbf{Exemplar 2}}:  The {\color{mypurple} \textbf{sales of new homes in the U.S. grow}} strongly.\\
					{\color{myblue}\textbf{Output 2}}: {\color{myyellow} \textbf{\emph{Sales of new homes in US}}} rise in May. \\		
					
					{\color{myblue}\textbf{Exemplar 3}}:  {\color{mypurple} \textbf{U.S. economic statistics: new home sales grow}} by 2.5 percent.\\
					{\color{myblue}\textbf{Output 3}}: {\color{myyellow} \textbf{\emph{US new home sales grow 26}}} percent in May. 
					
				\end{small}
			\end{minipage}
		}
	\caption{Two randomly sampled Gigaword development instances used for qualitative evaluation~(\S\ref{sec:analysis}).
		{\fontfamily{qcr}\selectfont Exemplar 1}'s are retrieved by the system (\S\ref{subsec:model:retrieval}),
		while the remaining ones are produced by the authors.
		Notable exemplars changes are highlighted in {\color{mypurple}\textbf{bold purple}},
		and output changes in {\color{myyellow}\textbf{\emph{italic yellow}}}.
		}
		\vspace{-.25cm}
	\label{fig:analysis}
\end{figure}

\section{Related Work}
\label{sec:related}
\paragraph{Exemplar-based generation.}
Partly inspired by traditional template-based generation~\interalia{kukich1983design,reiter1997building},
many recent efforts have been devoted to 
augmenting text generation models
with retrieved exemplars~\interalia{hodosh2013framing,mason2014domain,song2015two,lin2017adversarial}.
Without committing to an explicit slot-filling process, 
a typical method is to include exemplars as additional inputs to
the sequence-to-sequence models~\interalia{gu2017search,pandey2018exemplar,guu2018generating}.
\citet{wiseman2018learning} took a different approach and 
used a semi-Markov model to learn templates.
\paragraph{Dynamic parameter construction.}
The idea of using a smaller network to generate weights for a larger one
dues back to \citet{stanley2009hypercube} and \citet{koutnik2010evoling},
mainly under the evolution computing context.
It is later revisited with representation learning~\interalia{moczulski2015acdc,fernando2016convolution,alshedivat2017contextual},
and successfully applied to classification~\citep{xu2016dynamic,liu2017dynamic} 
and machine translation~\citep{platanios2018contextual}.
It also relates to the meta-learning set-up~\citep{thrun1998learning}.

\section{Conclusion}
\label{sec:conclusion}
We presented a text generation model
using exemplar-informed adaptive decoding. 
It reparameterizes the decoder using
the information gathered from retrieved exemplars.
We experimented with text summarization
and data-to-text generation,
and showed that the proposed model 
achieves strong performance and outperforms comparable baselines
on both.
The proposed model can be applicable in other conditioned 
text generation tasks.
We release our implementation at~\repo.

\section*{Acknowledgments}
We thank
Antonios Anastasopoulos,
Ming-Wei Chang,
Michael Collins,
Jacob Devlin,
Yichen Gong,
Luheng He,
Kenton Lee,
Dianqi Li,
Zhouhan Lin,
Slav Petrov,
Oscar T{\"a}ckstr{\"o}m,
Kristina Toutanova,
and other members of the Google AI language team
for the helpful discussion,
and the anonymous reviewers for their valuable feedback.

\bibliography{naacl2019}
\bibliographystyle{acl_natbib}
\clearpage

\begin{appendices}
\section{Implementation Details}

Our implementation is based on TensorFlow.\footnote{\url{https://www.tensorflow.org/}}
For both experiments,
we use the similar implementation strategies for the baselines and our model,
aiming for a fair comparison.
\subsection{Text Summarization}
We train the models using Adam~\citep{Kingma:2014} 
with a batch size of 64. 
We use the default values
in TensorFlow Adam implementation for initial learning rate $\eta_0$, $\beta_1$, $\beta_2$, and $\varepsilon$.
The models are trained for up to 20 epochs, 
with the learning rate annealed at a rate of 
0.2 every 4 epochs.
A weight decay of $0.01\times\eta$ is applied to all parameters,
with $\eta$ being the current learning rate.
The $\ell_2$-norms of gradients are clipped to 1.0.
Early stopping is applied based on \rouge-L performance on the development set. 

The weights of the output $\softmax$ function are tied with the word embeddings,
which are randomly initialized.
For the encoders,
we use 256-dimensional 3-layer BiLSTMs
in the Gigaword experiment,
and 300-dimensional 2-layer BiLSTMs
in the NYT experiment,
both with residual connections~\citep{he2015deep};
the decoders are one-layer (adaptive) LSTMs,
and have the same size as the encoders,
and so do the word embeddings.
We apply variational dropout~\citep{kingma2015variaonal} in the encoder RNNs,
and dropout~\citep{srivastava2014dropout} in the embeddings and the $\softmax$ layer,
the rates of which are empirically selected from $[0.15, 0.25, 0.35]$.
The last hidden state at the top layer of the encoder 
is fed through an one-layer $\tanh$-MLP,
and then used to as the decoder's initial state.
We use the attention function by \citet{luong2015effective},
and copy mechanism by \citet{see2017get}.
The exemplar encoder~(\S\ref{subsec:model:retrieval})
uses one-layer 32/50 BiLSTM for Gigaword and NYT experiments, respectively.
For numerical stability, $\bm{\lambda}$ (Equation~\ref{eq:lambda}) is
scaled to have $\ell_2$ norms of $\sqrt{d}$, with $d$ being the 
hidden size of the adaptive decoder~\citep{peng2018backprop}.

In the Gigaword experiment, we use 25K BPE types,
and limit the maximum decoding length to be 50 subword units;
while for NYT, we use 10K types with a maximum decoding length of 300,
and further truncate the source to the first 1000 units.
During evaluation, we apply beam search of width 5,
with a 1.0 length penalty~\citep{wu2016google}.

\subsection{Data-to-text Generation}
We in general follow the implementation details in the summarization experiment,
with the following modifications:
\begin{compactitem}
	\item We do \emph{not} apply byte-paired encoding here, and use a vocabulary of size 50K.
	\item The word embeddings are initialized using 840B version 300-dimensional \texttt{GloVe}~\citep{pennington2014glove}
	and fixed during training. Further, the $\softmax$ weights are \emph{not} tied to the embeddings.
	\item Three-layer 300-dimensional BiLSTM encoders are used, with residual connections; the exemplar encoder
	uses an one-layer 50-dimensional BiLSTM.
	\item Early stopping is applied based on development set \rouge-4 performance.
	\item A maximum decoding length of 40 tokens is used.
\end{compactitem}

\end{appendices}

\end{document}